\pdfoutput=1

\documentclass[11pt]{article}

\usepackage[]{acl}

\usepackage{times}
\usepackage{latexsym}

\usepackage[T1]{fontenc}

\usepackage[utf8]{inputenc}

\usepackage[nopatch]{microtype}  %

\usepackage{inconsolata}

\usepackage{booktabs,multirow}
\usepackage{colortbl}
\newcolumntype{C}[1]{>{\centering\let\newline\\\arraybackslash\hspace{0pt}}m{#1}}

\usepackage{arydshln}
\makeatletter
\def\adl@drawiv#1#2#3{%
        \hskip.5\tabcolsep
        \xleaders#3{#2.5\@tempdimb #1{1}#2.5\@tempdimb}%
                #2\z@ plus1fil minus1fil\relax
        \hskip.5\tabcolsep}
\newcommand{\cdashlinelr}[1]{%
  \noalign{\vskip\aboverulesep
           \global\let\@dashdrawstore\adl@draw
           \global\let\adl@draw\adl@drawiv}
  \cdashline{#1}
  \noalign{\global\let\adl@draw\@dashdrawstore
           \vskip\belowrulesep}}
\makeatother

\usepackage{graphicx}
\usepackage{subcaption}

\usepackage{setspace}

\newcommand{\repo}{https://github.com/cogstates/wikiface}

\newcommand{\tbf}[1]{\textbf{#1}}
\newcommand{\tss}[1]{\textsuperscript{#1}}

\usepackage{xspace}
\newcommand{\flabel}[1]{\textsc{#1}\xspace}

\newcommand{\SPosRBL}{\flabel{SPos+}}
\newcommand{\SPosTBL}{\flabel{SPos-}}
\newcommand{\HPosRBL}{\flabel{HPos+}}
\newcommand{\HPosTBL}{\flabel{HPos-}}
\newcommand{\SNegRBL}{\flabel{SNeg+}}
\newcommand{\SNegTBL}{\flabel{SNeg-}}
\newcommand{\HNegRBL}{\flabel{HNeg+}}
\newcommand{\HNegTBL}{\flabel{HNeg-}}

\newcommand{\Agreement}{\flabel{Agreement}}
\newcommand{\Confidence}{\flabel{Confidence}}
\newcommand{\Imposition}{\flabel{Imposition}}
\newcommand{\Disagreement}{\flabel{Disagreement}}
\newcommand{\Permissiveness}{\flabel{Permissiveness}}
\newcommand{\Autonomy}{\flabel{Autonomy}}
\newcommand{\Apologies}{\flabel{Apologies}}
\newcommand{\Indebtedness}{\flabel{Indebtedness}}

\newcommand{\SPosR}{\Confidence}
\newcommand{\SPosT}{\Apologies}
\newcommand{\HPosR}{\Agreement}
\newcommand{\HPosT}{\Disagreement}
\newcommand{\SNegR}{\Autonomy}
\newcommand{\SNegT}{\Indebtedness}
\newcommand{\HNegR}{\Permissiveness}
\newcommand{\HNegT}{\Imposition}
\newcommand{\Other}{\flabel{None}}

\newcommand{\sddag}{\tss{\textdaggerdbl}}

\newcommand{\bandlp}{B\&L\xspace}
\newcommand{\bandlt}{B\&L\xspace}

\usepackage{enumitem}
\setlist[enumerate]{
    topsep=0pt,itemsep=-0.5ex,
    partopsep=0.5ex,parsep=0.5ex
}
\setlist[itemize]{
    topsep=0pt,itemsep=-0.5ex,
    partopsep=0.5ex,parsep=0.5ex,
}

\usepackage{fontawesome5}
\newcommand{\DoCS}{{\tiny\faLaptop}}
\newcommand{\DoL}{{\tiny\faComments}}
\newcommand{\IACS}{{\tiny\faCalculator}}

\newcommand{\secref}[1]{\S\ref{#1}}

\usepackage{todonotes}

\definecolor{ocr}{HTML}{009900}
\definecolor{shynecolor}{HTML}{33FFD7}

\definecolor{as}{HTML}{0000FF}

\title{Examining Gender and Power on Wikipedia Through Face and Politeness}

\author{
Adil Soubki\tss{\DoCS\hspace{0.1em}\IACS}, Shyne Choi\tss{\DoCS}, Owen Rambow\tss{\DoL\hspace{0.15em}\IACS} \\
\tss{\DoCS}Department of Computer Science, \tss{\DoL}Department of Linguistics \\
\tss{\IACS}Institute for Advanced Computational Science, Stony Brook University \\
\texttt{asoubki@cs.stonybrook.edu,}
\texttt{\{shyne.choi,owen.rambow\}@stonybrook.edu}
}

\begin{document}
\maketitle
\begin{abstract}

We propose a framework for analyzing discourse by combining two interdependent concepts from sociolinguistic theory: face acts and politeness. While politeness has robust existing tools and data, face acts are less resourced. We introduce a new corpus created by annotating Wikipedia talk pages with face acts and we use this to train a face act tagger. We then employ our framework to study how face and politeness interact with gender and power in discussions between Wikipedia editors. Among other findings, we observe that female Wikipedians are not only more polite, which is consistent with prior studies, but that this difference corresponds with significantly more language directed at humbling aspects of their own face. Interestingly, the distinction nearly vanishes once limiting to editors with administrative power.

\end{abstract}

\section{Introduction} \label{sec:intro}

\noindent
\citet{brown-levinson-1987-politeness} (henceforth \bandlt) introduce an influential theory of politeness based on the concept of face, which they claim to be culturally universal. In this theory, face -- i.e. the public image one seeks to claim -- is a two-sided coin. Agents attend to their desire to have their wants appreciated, which they call positive face, as well as a complementary desire to act unimpeded and maintain freedom, which they call negative face. The face of every agent is ensnared with that of every other agent -- agents cannot have their desires appreciated if they cannot appreciate the desires of others. As a result, utterances can raise (+) or threaten (-) the positive (Pos) or negative (Neg) face of the speaker (S) or hearer (H).

A face threat or face raising is not a property of particular linguistic choices, but of communicative intent.
If I want to request information from you, then I necessarily need to threaten your negative face, since, if I am successful in communicating my request to you, I will oblige you to answer and thus I will restrict your choice of actions.
In \bandlt's theory, discourse participants may choose among various strategies for minimizing threats to face.
These strategies are linguistic strategies (for example, using hedges), and the choice of strategy depends on many factors such as cultural conventions and the discourse situation (who is talking to whom under what circumstances).

Work related to NLP has concentrated on studying linguistic manifestations of politeness \cite{walker-etal-1997-improvising,danescu-niculescu-mizil-etal-2013-computational} while largely disregarding the notion of face act.
While \bandlt are frequently cited, the deep insight of their theory comes from a complexity which has been ignored. Their theory is not simply about politeness, but about how politeness, situated in the context of rational action, manifests from a combination of performing face acts to achieve certain goals and using mitigation strategies to lessen the impact of face-threatening acts. \citet{danescu-niculescu-mizil-etal-2013-computational} use politeness markers inspired by \bandlt strategies as features for a system which predicts perceived politeness without modeling face acts. \citet{dutt-etal-2020-keeping} predict face acts in isolation from perceived politeness. In this paper, we re-examine the Wikipedia Talk Pages Corpus \cite{danescu-niculescu-mizil-etal-2011-echoes,chang-etal-2020-convokit} and demonstrate how bringing face acts and politeness together provides deeper insight. %

We do this by producing an annotation of face acts on the corpus and training a new model to label utterances. We then use this tool, along with prior systems which produce judgements of perceived politeness, to label roughly 1.3 million sentences from Wikipedia talk pages. To our knowledge, we are the first to apply an annotation grounded in politeness theory to a text corpus of this scale.

The paper is structured as follows. We start with a review of relevant literature (\secref{sec:related}) and present our theoretical framework (\secref{sec:theoretical-framework}).
We then turn to producing an annotation of face acts on the Wikipedia Talk Pages Corpus and building a tagger using this new dataset (\secref{sec:fa-wiki}). Our framework is then applied by bringing this new tagger together with existing tools to re-analyze the corpus, paying special attention to gender and power (\secref{sec:application}). We end by reporting our conclusions along with a discussion of future work (\secref{sec:conclusion}).

All of the code written, datasets prepared, and experimental observations made in the course of this research will be made available on \href{\repo}{GitHub}.\footnote{\url{\repo}}

\begin{table*}[!ht]
\centering
\resizebox{0.8\linewidth}{!}{
\begin{tabular}{lll}
\toprule
\bfseries Face Act & \bfseries Mnemonic & \bfseries Sample Discourse Goals \\
\midrule
\HNegTBL & \Imposition & Requests, commands, questions, offers, promises, ... \\ 
\HPosTBL & \Disagreement & Criticism, insults, disapproval, ... \\
\HNegRBL & \Permissiveness & Granting permission, making exceptions, ... \\
\HPosRBL & \Agreement & Seeking common ground, group cohesion, \ldots \\
\midrule
\SNegTBL & \Indebtedness & Thanking, accepting offers or thanks, commitments, ... \\
\SPosTBL & \Apologies & Confessions, embarrassment, ... \\
\SNegRBL & \Autonomy & Refusing requests, asserting freedoms, ... \\
\SPosRBL & \Confidence & Self-promotion, signaling virtue, ... \\
\bottomrule
\end{tabular}
}
\caption{Face acts with mnemonic label and examples of discourse goals.}
\label{tab:nicknames}
\end{table*}

\section{Related Work}
\label{sec:related}

The theory of politeness of B\&L has found applications in many  fields including sociology, psychology, and linguistics.  Google Scholar lists nearly 38,000 citations.  Curiously, in NLP there has not been 
much work building explicitly on B\&L. \citet{danescu-niculescu-mizil-etal-2013-computational} concentrate on one type of face-threatening act (FTA), namely the negative face-threatening act of a request, and investigate the strategies used for doing this FTA. To do this, they use crowd sourcing to rate the requests on a politeness scale. They develop a model which predicts the politeness of these requests and use it to study the interactions between users on Wikipedia and StackExchange.  \citet{ziems2023css} show that fine-tuning on the data of \citet{danescu-niculescu-mizil-etal-2013-computational} substantially outperforms zero-shot approaches.

The face acts (FAs) themselves are the object of \citet{dutt-etal-2020-keeping}. In addition to developing a dataset annotated with FAs, they present a FA classifier based on a neural architecture they devise on top of BERT, which achieves 69\% F-measure (0.60 macro). As the data involves participants convincing others to donate to a charity, they also use this corpus to investigate the relationship between face acts and persuasion by predicting if a participant chose to donate.  This corpus, which we refer to as the ``CMU Face Acts Corpus'' (or ``CMU Corpus'' for short) in this paper, is the direct inspiration for our annotation effort on the Wikipedia data.  We differ from their annotation scheme in some important details; we present our annotation in \secref{sec:fa-wiki}. In prior work, we investigated the interaction of intention (through dialog act tagging) and face acts in the CMU Corpus \citep{soubki-rambow-2024-intention}.

There has been an explosion work in computational social science in general, in which NLP tools are used to extract relevant signals from large amounts of data in order to study a social phenomenon, such as changing attitudes towards certain topics as expressed on social media.  For an overview, see \citep{annurev-soc-121919-054621}.  In the area of studying how gender and power shape written dialogs, there has been some work in NLP.  
Working with corporate emails, \citep{prabhakaran-reid-rambow:2014:EMNLP2014} find that gender differences become exaggerated when looking at individuals with greater social power; specifically, among people with power, women behave {\em more} differently from men than when comparing people without power.

Finally, turning to the study of politeness and gender outside of NLP, there have been some studies based on manual analysis of collected data, for example \cite{herring-1994-politeness,tannen:1994a,kunsmann:2013}.  
For space reasons, we discuss only one example in more detail. \citet{kendall:2005}, using a framing approach following \cite{goffman:1974}, finds that women in power who ``downplay status differences (...) are exercising and constituting their authority by speaking in ways that accomplish work-related goals while maintaining the faces of their interlocutors''.  In the terminology of B\&L (which \citet{kendall:2005} does not use), women perform similar face acts to men but use strategies to mitigate the effects, which results in women in power appearing more polite than men in power.

\section{Theoretical Framework} 
\label{sec:theoretical-framework}

In this section we provide a brief summary of relevant concepts from politeness theory as it relates to our work.
Our goal in this paper is to explore how face acts contribute to the perception of politeness.   For \bandlt, ``face'' refers to the public self-image of agents, and it is a universal component of human interaction.  It consists of two complementary facets \citep[\S3.1, p. 61]{brown-levinson-1987-politeness}.
(1) negative face: ``the basic claim to territories, personal preserves, rights to non-distraction -- i.e. to freedom of action and freedom from imposition.''
(2) positive face: ``the positive consistent self-image or `personality' (crucially including the desire that this self-image be appreciated and approved of) claimed by interactants.''

A face act is an intentional communicative act
which inherently interacts with the face of the speaker and/or addressees \citep[\S3.2, p. 65]{brown-levinson-1987-politeness}.  Face acts can threaten (-) or affirm (+) the face; they can be about the speaker's face (S) or the hearer's (H); and they can be about positive (Pos) or negative (Neg) face.  This gives us eight possible face acts, shown in Table~\ref{tab:nicknames}, where we also provide a short mnemonic names which we will use in this paper, as the terminology of \bandlt can be unintuitive.  

Face acts are part of a larger sequence of choices a speaker makes.  First, the speaker chooses a discourse goal or goals (which may form a hierarchy) which will be realized in a speech act \citep{austin:1962}; then they determine which face acts contribute to the discourse goals; they then choose a strategy to realize this face act, in conformance with the cultural norms of their community which are mutually known by them and the hearer in the communicative context (age, gender, power differential of the discourse participants); and finally, they produce the utterance, which the hearer will perceive as more or less polite, given the discourse goal of the speaker, the communicative context, and the mutually known cultural norms.  We see that the notion of ``strategy'' plays a crucial role in the mediation between face act performance and perceived politeness, and \bandlt devote a large portion of their study to strategies.  Unfortunately, there are no corpora annotated for face act strategies.\footnote{\citet{danescu-niculescu-mizil-etal-2011-echoes} use a notion of ``strategy'' which is defined by a grouping of lexical items that are assumed to affect the hearer's perception of politeness.  They 
can be considered a simple approximation of the notion in \bandlp, and in fact helps in predicting politeness.  We have chosen not to use these ``stratgeies'' (though they are straightforward to determine, as they are based exclusively on word matching), since we would like to address the issue in a more principled manner in the  future.}

We emphasize that face acts do not imply perceived politeness (\secref{sec:appendix-corrs}). Consider the following examples from the Wikipedia corpus.

{\addtolength\leftmargini{-0.1in}
\begin{quote}
\small
\ttfamily
\noindent
\hspace{-1em}[\tbf{1}]
\noindent
B: Why open a peer review when we are looking for someone to do the GA review? \\
\textcolor{white}{.\hspace{0.2em}.}A: Why request a second GA, 3 days after the first one failed?

\hspace{-1em}[\tbf{2}]
A: Hi Plange, any reason why this category is named differently to the others?
\end{quote}}

\noindent Both utterances are \HNegTBL/\HNegT face acts, because they impose on the hearer the obligation to respond.  However, (1) rejects the previous question by B and challenges B, while (2) is just a request for information, so that (1) is perceived as more impolite than (2).

It is possible for a single utterance to perform multiple face acts at once. For example, (1) could also be seen as \HPosT, since it entails a critique of B's actions.
However, \citet{dutt-etal-2020-keeping} observed multi-labeled acts in only 2\% of their data, leading them to consider a single label per utterance. We make this simplification as well in the work presented in this paper.

\section{Face Act Tagging}
\label{sec:fa-wiki}

In this section we outline the data, modeling techniques, and evaluation measures used in developing our face act tagger for Wikipedia talk pages.

\subsection{Dataset}
On Wikipedia, talk pages are used by editors to coordinate changes and improvements to the encyclopedia.\footnote{\url{https://en.wikipedia.org/wiki/Wikipedia:Talk_page_guidelines}} 
A variety of social and power dynamics are at play in these conversations which can range from discussions of bureaucratic process to heated, and sometimes personal, conflicts. The Wikipedia Talk Pages Corpus \citep{danescu-niculescu-mizil-etal-2011-echoes} collects 125,292 exchanges between 38,462 editors resulting in a total of 391,294 posts for analysis. Unlike the CMU Face Acts Corpus, where participants are on mostly level ground, editors can hold administrative privileges or greater notoriety within the community, resulting in interactions with large social distance. 
Additionally, some editors self-identify gender on their user page.\footnote{\url{https://en.wikipedia.org/wiki/Wikipedia:User_pages}}
This is desirable in our case as it allows us to study how these social factors interact with face and politeness.

There can be nested replies in talk pages which allow for situations where an utterance is not a reply to the preceding utterance. We do not attempt to correct for these cases and sort first to preserve reply structure and then by the time of the post.

\subsection{Annotation} \label{sec:annotation}

Similar to the CMU Corpus, we use the criteria outlined by \bandlt, which serves as our reference. The CMU Corpus annotation guidelines, as the authors noted, contain some departures from politeness theory. In particular, the CMU Corpus annotates both thanking and complimenting as \HPosR .  In contrast, \bandlt analyze thanking and complimenting as \SNegT and \HNegT, respectively. We choose to remain faithful to \bandlt,
and in fact assert this to be a critical piece of the theory.  Consider a compliment such as \emph{you have a lovely smile}.  How is it that a compliment can be taken so poorly by the addressee if the speaker is not risking anything? They are often very risky social acts because the speaker assumes they are among the people their addressee wishes to be complimented by; a very imposing assumption. Thanking, on the other hand, can be seen as an exchange of currency. Similar to writing an IOU, the speaker offers a token of their freedom to the addressee.
We note that we expect future versions of face act annotations to annotate multiple face acts at once, which may resolve this difference between the CMU Corpus annotation style and ours.

We randomly selected 200 conversations from the WikiTalks data for manual annotation. As the posts contain multiple sentences, each with the possibility of their own face act, we segment the sentences prior to annotation using spaCy \citep{honnibal-johnson-2015-improved}. To reduce errors in segmentation, we scrubbed hypertext tags and masked any remaining urls. This resulted in 1850 sentences.  We will refer to these basic units of annotation as ``utterances'' in the following sections.  Two of the authors annotated the 1850 utterances for face acts. We examined 100 utterances labeled by both annotators and computed a Cohen's Kappa of 0.69 which indicates moderate to substantial agreement.

\subsection{Modeling}
We model face act tagging as a text classification task. Given a sequence of $n$ utterances $S = [t_1, t_2, \dots, t_n]$, we wish to assign a label $y \in Y$ where $Y$ represents a set containing the 8 possible face acts and one additional label for no face act.
Recently, many classification tasks have achieved stronger results using parameter efficient fine-tuning methods of larger models rather than full fine-tuning smaller ones \citep{hu-etal-2022-lora,dettmers-etal-2024-qlora}. We adopt this approach and use Llama-3-8B \citep{dubey-etal-2024-llama3} and LoRA with Int8 quantization \citep{dettmers-etal-2022-int8} for fine-tuning.\footnote{Our choice of Llama-3 was informed by a preliminary set of experiments in which a variety of pre-trained models and methods were were examined on single seed runs.}
Details of the configuration are given in Appendix~\ref{sec:appendix-config}.

\subsection{Data Representation} \label{sec:data-representation}

While fine-tuning approaches unify many aspects of the model design, they present challenges when it comes to determining effective input and output representations.

We provide the models an input which contains an utterance prefixed with the Wikipedia username of the discourse participants,\footnote{We note that the Wikipedia usernames shield the actual identity of the discourse participant, and that the Wikipedia username is public.} along with previous utterances as context.  
Each utterance is followed by a newline character. %
We give an example with two lines of context, though in our experiments we use more, as discussed just below.

{\addtolength\leftmargini{-0.1in}
\begin{quote}
\small
\ttfamily
\noindent
\hspace{-1em}[\tbf{Input}] \\
\noindent
Jossi: I will. \\
Jossi: Just play nice, that is all I ask. \\
Kelly: What's that supposed to mean?

\hspace{-1em}[\tbf{Output}] \\
hpos- 
\end{quote}}

The target output is 
a distribution where the highest probability is given to the correct label %
for the final utterance of the input text, in this case \HPosTBL (\HPosT). We experimented with different output formats, and found they do not make much of a difference.  In our experiments we noticed context to be a critical factor with the optimal size varying by model. Llama 3 performed best with a size of four, for a total of five utterances. As there are no previous turns for the first four turns in each dialog, those examples are provided in a similar format containing only three, two, one or no lines of context.

\begin{figure*}[!ht]
    \centering
    \includegraphics[scale=0.4]{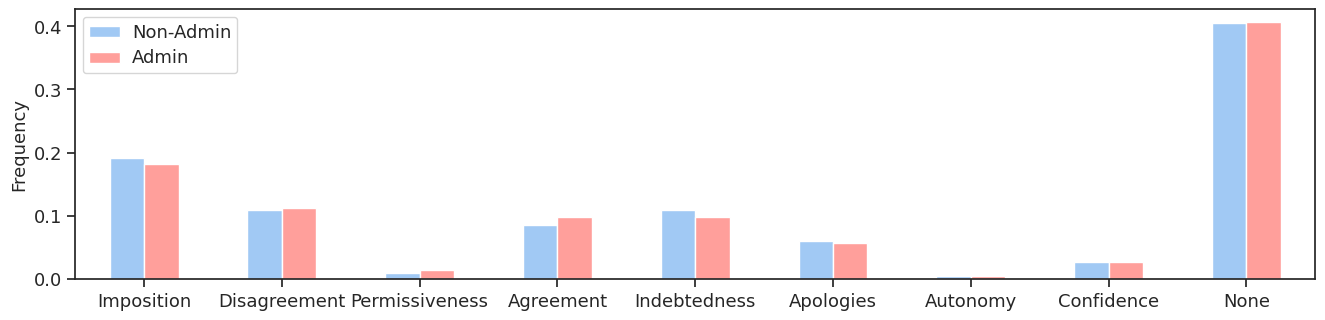}
    \caption{Frequency of face acts for admins and non-admins.}
    \label{fig:fas-vs-admin-status}
\end{figure*}

\subsection{Experimental Setup And Evaluation} \label{sec:approach-eval}

We perform all experiments using five-fold cross validation and the evaluation metrics are averaged across all five folds.
We evaluate model performance using F-measure for each of the nine classes as well as micro and macro F-measure aggregated over all labels.
We performed hyperparameter tuning, and report metrics only for the best model.

\begin{table}[!b]
\centering
\begin{tabular}{lc}
\toprule
\bfseries Micro & 0.68  \\
\bfseries Macro & 0.51  \\
\midrule
\bfseries \HNegT & 0.73  \\
\bfseries \HPosT & 0.56  \\
\bfseries \HNegR & 0.40  \\
\bfseries \HPosR & 0.58  \\
\midrule
\bfseries \SNegT & 0.80  \\
\bfseries \SPosT & 0.56  \\
\bfseries \SNegR & 0.04  \\ %
\bfseries \SPosR & 0.14  \\ %
\midrule
\bfseries \Other & 0.76  \\
\bottomrule
\end{tabular}
\caption{Mean F1 across all folds of our annotation.}
\label{tab:wiki-results}
\end{table}

\subsection{Results}

The results of these experiments are reported in Table~\ref{tab:wiki-results}.
We achieve a micro-averaged F1 of 0.68 (average across five folds). %
Since the task is, with the exception of some nuances (\secref{sec:annotation}), identical to the CMU Face Acts Corpus we also tried continued training on the CMU Face Acts Corpus, but this did not improve performance.
We suspect this is due to the difference in genre and slight change in annotation procedure, which results in a different distribution of labels between the two datasets.

\section{Application and Analysis} \label{sec:application}

\noindent
We apply our new face act tagger along with the politeness scores %
provided by ConvoKit \citep{chang-etal-2020-convokit} to study the interactions of face and politeness over the entire Wikipedia Talk Pages Corpus. Our face act tagger is trained using our entire annotation (\secref{sec:annotation})
before applying it to the Wikipedia data. This produces roughly 1.3 million sentences labeled with face acts and perceived politeness.
We note that the politeness scores are obtained for the entire turn, as this is what the perceived politeness model is trained on, while face acts are tagged by sentence to allow for greater granularity.

In our analysis of politeness we investigate how polite (magnitude) editors are perceived to be by looking at their scores and how often that occurs (frequency) by considering the proportion of utterances in the top 25\% of politeness scores. For face acts, we compare the overall distribution (frequency) of labels. Statistical significance is calculated using the Mann-Whitney U test. This analysis was also performed on only the human annotated portion of the data and the trends remained consistent. We report results on the entire corpus.

\begin{figure*}[!t]
    \centering
    \includegraphics[scale=0.4]{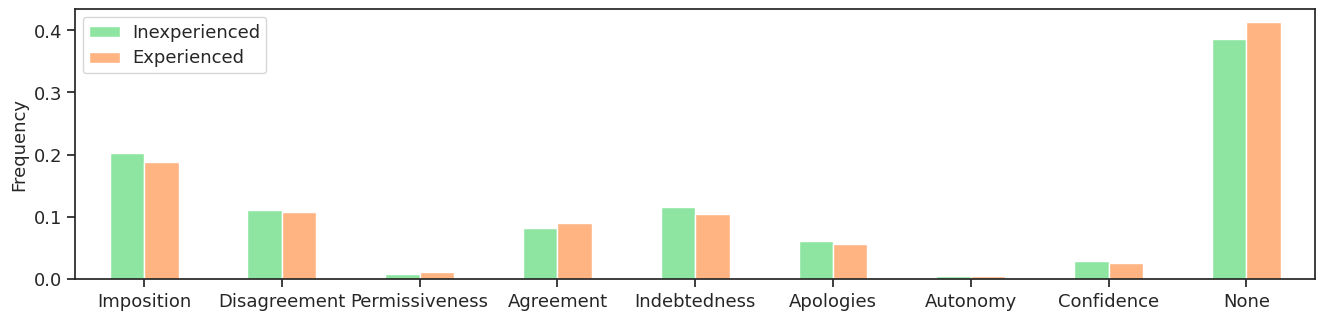}
    \caption{Frequency of face acts by editor experience}
    \label{fig:exp-vs-fas}
\end{figure*}

\noindent

\subsection{Admin Differences} \label{sec:admin-diffs}

On Wikipedia, editors with administrative status wield significant power in the community including the ability to block or unblock users by IP address and delete or restore pages.
This increased status is known to be recognized in the community \citep{danescu-niculescu-mizil-etal-2011-echoes,burke-kraut-2008-taking,leskovec-etal-2010-governance}
which endows editors with these powers through public elections.
We note that politeness theory anticipates speakers with greater social power than their addressee to more often select strategies that reduce ambiguity and lengthiness. This means opting to perform face threatening acts more often (as opposed to avoiding them all together) and mitigating them through the trade-offs of strategies less often, which one would expect to correspond with a perception of being less polite overall.

We divide utterances by their politeness score into the polite utterances (top 25\%), neutral (next 50\%) and impolite (bottom 25\%). When comparing politeness between admins and non-admins we see the same trend as observed by \citet{danescu-niculescu-mizil-etal-2013-computational}. Utterances produced by editors with administrative privileges (``admins'') are not more often impolite, however they are significantly ($p < 0.001$ using the Mann-Whitney U test) less frequently polite, with a mean score difference of 3.
Additionally the frequency by which admins produce polite posts is also significantly ($p < 0.001$) lower resulting in messages which are deemed polite 5\% less often compared to non-admin editors.

When looking at the distributional differences in face acts by adminship (Figure~\ref{fig:fas-vs-admin-status}) this decrease in politeness corresponds with small, but salient variations. Admins are significantly ($p < 0.001$) less likely to express \SNegT (e.g. thanking, accepting offers) and \SPosT (e.g., admitting mistakes, confessions). Though admins produce more utterances labeled \HPosR (e.g. appreciation, seeking common ground, group cohesion), their \HPosR utterances are significantly ($p < 0.001$) less often ($-4\%$ absolute) perceived as polite compared to \HPosR utterances by non-admins.
Similarly, while non-admins do more \HNegT (e.g. issuing commands, making requests), their \HNegT utterances are significantly ($p < 0.05$) more often ($+3\%$ absolute) taken politely compared to \HNegT utterances by admins..  This shows, as we anticipated, that face acts do not imply politeness, contrary to possible intuition.\looseness=-1

\subsection{Experience Differences}

\begin{table}[t]
\centering
\begin{tabular}{lr}
\toprule
 & \bfseries Politeness \\
\midrule
\bfseries Experienced Admin & 0.34\tss{$\dagger$} \\
\bfseries Experienced Non-Admin & 0.36\tss{$\dagger$}\\
\bfseries Inexperienced Admin & 0.38\tss{$\dagger$} \\
\bfseries Inexperienced Non-Admin & 0.40\tss{$\dagger$} \\
\bottomrule
\end{tabular}
\caption{Mean politeness scores for difference admin types. All differences are found to be significant using the Mann-Whitney U test with $p < 0.001$.}
\label{tab:exp-x-admin-vs-politeness}
\end{table}

We explore whether the experience and productivity of the editor is another means to achieve increased social power without the explicit additional privileges the ``admin'' title confers. To investigate this we categorize users by the number of edits they have made and label users in the top and bottom quartiles ``experienced'' and ``inexperienced'', respectively. %

\begin{table}[!tb]
\centering
\begin{tabular}{lrrr}
\toprule
& \bfseries Inexperienced & \bfseries Experienced \\
\midrule
\bfseries Impolite & 0.07 & 0.07 \\
\bfseries Polite & 0.35\tss{$\ddagger$} & 0.28\tss{$\ddagger$} \\
\bottomrule
\end{tabular}
\caption{Proportion of turns classified as (im)polite by editor experience level. $\ddagger$ indicates significance with $p < 0.0001$ using the Mann-Whitney U test.}
\label{tab:exp-vs-politeness}
\end{table}

\begin{figure*}[ht]
    \centering
    \includegraphics[scale=0.4]{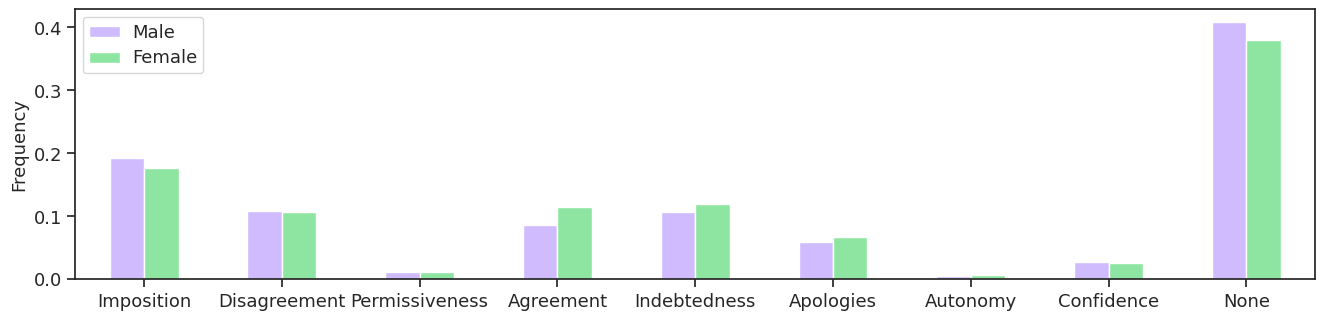}
    \caption{Frequency of face acts by gender.}
    \label{fig:fas-gender}
\end{figure*}

We observe similar trends in politeness among experienced editors (Table~\ref{tab:exp-vs-politeness}) to that of admins, with turns by experienced editors being labeled polite 7\% less often relative to inexperienced editors.
When looking at the differences in face acts (Figure~\ref{fig:exp-vs-fas}) we note that there are ways in which new-comers behave like experienced Wikipedians such as a willingness to the face act \HPosT. However, like admins, experienced users are significantly ($p < 0.001$) less likely to express \SNegT or \SPosT. Unlike when comparing by admin status, we find that experienced admins are significantly ($p < 0.001$) less likely to interact with face all together (more labeled \Other).

We now investigate how experience interacts with admin status. As expected, experience is correlated ($r=0.37$) with adminship with nearly half of all admins landing in the top quartile of editors by edit count. 
We find admins in the top quartile by edit count are significantly (p < 0.001) less polite than the bottom quartile.
Additionally, intersecting experience with admin status (Table~\ref{tab:exp-x-admin-vs-politeness}) finds a spectrum. Experienced admins are the least polite but experienced non-admins are less polite than inexperienced admins. 
This indicates that these factors are additive in their contribution to social power.

\subsection{Gender Differences} \label{sec:gender-diffs}

Some editors self-identify their gender on their user page
allowing us to study communicative differences along this axis as well.
Prior work found female Wikipedians to be generally more polite \citep{danescu-niculescu-mizil-etal-2013-computational} which is consistent with studies in several domains. We also observe this, with utterances by women scoring more polite (+5, $p < 0.001$), more often (+7\%, $p < 0.0001$).

When comparing the distribution of face acts (Figure~\ref{fig:fas-gender}) we see several disparities that the politeness scores alone do not convey. In general, the \Other category is lower for women, i.e. female Wikipedians are more likely, and perhaps more willing, to interact with face in their utterances.  When doing so, they humble their own positive face (\SPosT, e.g. admitting mistakes, making confessions, accepting compliments) and their own negative face  (\SNegT, e.g. thanking, accepting apologies) more often than men. This self-deference is accompanied by fewer impositions on their addressee's face (\HNegT, e.g. requests, commands, insults, criticism) and more attention to the hearer's own wants (\HPosR, e.g. seeking common ground, showing respect). Unlike when looking at admins, these \HPosR utterances are less frequently judged to be impolite. These trends have been observed in various prior studies \citep{lakoff-1973-language,prabhakaran-rambow-2017-dialog,herring-1994-politeness}.%

\subsection{Intersectional Differences}

We have seen that male Wikipedians are less polite, more distant with regards to face, and more likely to express \HNegT (\secref{sec:gender-diffs}).
Similarly, much of the same is true when comparing admins to non-admins (\secref{sec:admin-diffs}). How do these factors interact?  As mentioned in \secref{sec:related}, previous work in other domains has found gender differences to become exaggerated in the communication patterns of individuals with power. One might expect a similar trend to hold on Wikipedia. %

\begin{table}[tb]
\centering
\begin{tabular}{lrr}
\toprule
& \bfseries Male & \bfseries Female \\
\midrule
\bfseries Non-Admin\sddag & 0.37 & 0.43 \\
\bfseries Admin & 0.34 & 0.35 \\
\midrule
\bfseries Inexperienced\sddag & 0.41 & 0.43 \\
\bfseries Experienced\sddag & 0.34 & 0.42 \\
\bottomrule
\end{tabular}
\caption{Mean politeness scores by experience and admin status compared across gender. $\ddagger$ indicates significance with $p < 0.0001$ using the Mann-Whitney U test when comparing across gender.}
\label{tab:exp-admin-gender-politeness-score}
\end{table}

When comparing politeness across both gender and administrative status
(Table~\ref{tab:exp-admin-gender-politeness-score}), we find that this does not appear to be the case. 
While women admins are more polite (magnitude) than male admins, the difference is not significant ($p > 0.1$). 
Meanwhile, their non-admin counterparts are significantly more polite than non-admin men (+6, $p < 0.0001$).
Among non-admin editors, women produce utterances in the top quartile of politeness 10\% more often than men, while this reduces to just 1\% when comparing admins across genders.

\begin{figure}[!t]
    \centering
    \resizebox{0.85\linewidth}{!}{
    \includegraphics[width=\textwidth,height=16cm]{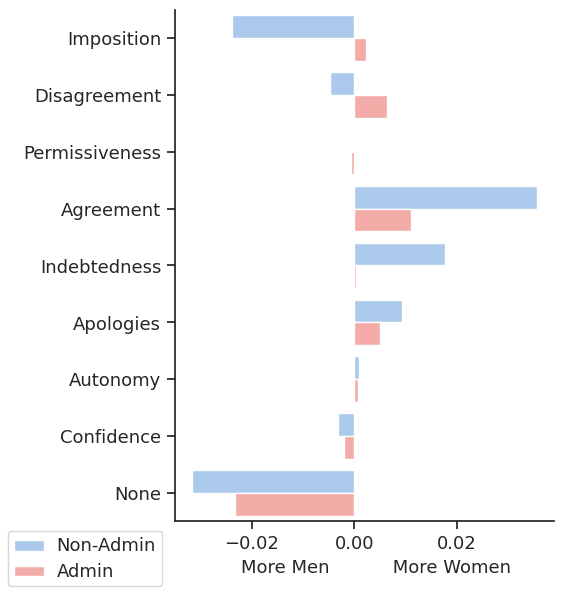}
    }
    \caption{Differences between relative usage of face acts by gender, broken down by non-admins (blue) and admins (red); lines to the right (left) indicate that women (men) perform the face act more often}
    \label{fig:gender-x-admin-vs-fas}
\end{figure}

Overall the distribution of face acts (Figure~\ref{fig:gender-x-admin-vs-fas}) between male and female admins is similar to that of non-admins (the red lines for admins and blue lines for non-admins in Figure~\ref{fig:gender-x-admin-vs-fas} are in the same direction), except that the difference between men and women is reduced (the red lines are shorter than the blue lines).%
There is one striking exceptions: 
among non-admins, men make many more \HNegT (e.g., making requests, issuing commands) 
face acts than women, but this difference disappears for admins (and in fact women perform \HNegT utterances slightly more frequently than men).
We note that {\HNegT} is the face act that becoming an admin specifically entitles the editor to perform: admins have the right to request changes (and that changes be undone). 
We speculate that female admins specifically make use of their
socially sanctioned power, while men perform \HNegT acts even when having no specific admin authority.
In summary, admin privileges maintain but substantially lessen the previously observed gender differences in politeness and face.
Put differently, female admins behave more like men (whether admins or not), which we also saw in the politeness scores (Table \ref{tab:exp-admin-gender-politeness-score}).

We now turn to the intersection of gender and experience.  Here, we see a strikingly different result.  
For all conditions (non-admin, admin, inexperienced, experienced), women are more polite.  However, 
we see from 
Table~\ref{tab:exp-admin-gender-politeness-score}
that men become more impolite as they become experienced, while this is not the case for women: there is no significant change in their politeness as they become experienced.  The only exception is for women who become admins (who are, often, experienced), who behave as men do.  
Put differently, experience and the official power designator of ``admin'' do not function in the same way across gender: for men, both result in less politeness, but for women, only the ``admin’’ title does.  

\begin{figure}[!t]
    \centering
    \resizebox{0.85\linewidth}{!}{
    \includegraphics[width=\textwidth,height=16cm]{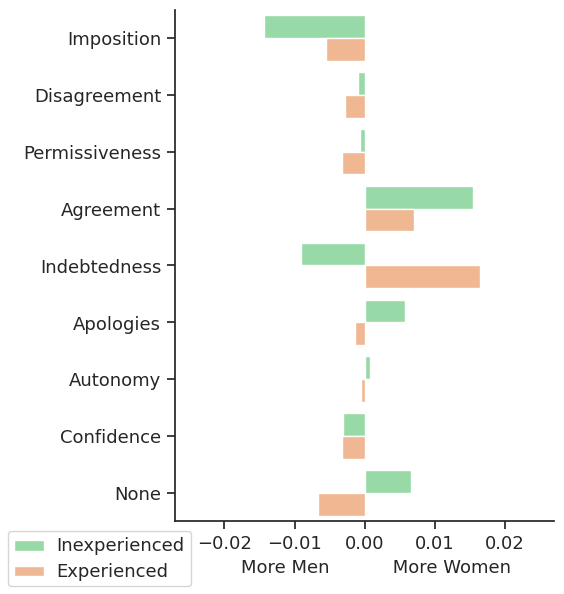}
    }
    \caption{Differences between relative usage of face acts by gender, broken down by inexperienced (green) and experienced (orange) users; lines to the right (left) indicate that women (men) perform the face act more often}
    \label{fig:gender-x-experience-vs-fas}
\end{figure}

When looking at face acts (Figure~\ref{fig:gender-x-experience-vs-fas}), we see that for some categories the differences between men and women are reduced with experience (the orange bars are shorter than the green bars). However, a notable exception is for \SNegT, for which we see a large increase in the difference between men and women, and in fact a flip in which gender performs it more often.  When looking at the absolute numbers (not shown in the table), we can see why: women do not change the frequency of their \SNegT utterances at all as they gain experience, while men decrease their frequency of \SNegT utterances from 12.3\% to 9.8\% of their utterances.  This decrease is a major contributor to the decrease in politeness among experienced men (but not among experienced women).  We extend our previous interpretation by speculating that experienced women do not feel they have a socially sanctioned position of power, and/or men experience a decrease in social distance towards other Wikipedians as they become more experienced, while women do not.

\section{Conclusion and Future Work}
\label{sec:conclusion}

We identify an optimized method for training face act taggers using fine-tuning on LLMs, contribute a new corpus annotated for face acts, and make available a pre-trained model for use on Wikipedia. Through several methods of analysis we demonstrate the usefulness of examining perceived politeness in combination with face acts by reporting a number of findings based on their interaction. In future work we plan to allow multiple face acts per utterance (including for the same segment), and to incorporate the strategy (as conceived of by \bandlt) more explicitly into our modeling framework.

\section*{Limitations}

The principal scientific limitation of this work is that we could only consider three aspects of the larger model of \bandlt: face acts, the communicative setting (gender and power), and perceived politeness.  The major missing elements in the full framework include intention, communicative intention, social norms, and strategies.  We intend this paper to be a first step towards a fuller implementation of an explicit cognitive theory of communication which involves all of the mentioned elements.

The experiments for this work were performed using computational resources that are not, in general, freely available. In part due to these computational requirements, but also a result of minimal data, we were not able to evaluate the techniques on additional languages and acknowledge the limitations this places on extending our results to other cultures. We also note along similar lines that while \citet{brown-levinson-1987-politeness} claim their theory of politeness to be culturally universal, this claim has been contested -- most notably for eastern cultures \citep{al-duleimi-etal-2016-critical}. As discussed in detail above, taking utterances to have a single face act or intent is a critically limiting assumption which lends some uncertainty to our conclusions.

We note that while many of the linguistic differences observed were consistent across multiple rounds of analysis and significant using the Mann-Whitney U test, the effect sizes were generally small. The conclusions should be interpreted with that in mind.

\section*{Ethics Statement}
Despite an analysis of the errors, we cannot verify the safety of this system in any user-oriented context and therefore do not recommend such uses without further study. While we do not produce any datasets directly from human annotations, we do use several datasets which were, to the best of our knowledge, compiled ethically. As the primary object of study in this work is the relationship between politeness and language, we do not anticipate broad risks to its application.

\section*{Acknowledgements}

This material is based upon work supported by the Defense Advanced Research Projects Agency (DARPA) under the CCU (No. HR001120C0037, PR No. HR0011154158, No. HR001122C0034) program.  
Soubki has received additional support from the National Science Foundation (NSF) under No. 2125295 (NRT-HDR: Detecting and Addressing Bias in Data, Humans, and Institutions).
Any opinions, findings and conclusions or recommendations expressed in this material are those of the author(s) and do not necessarily reflect the views of the NSF or DARPA.

We thank both the Institute for Advanced Computational Science and the Institute for AI-Driven Discovery and Innovation at Stony Brook for access to the computing resources needed for this work. These resources were made possible by NSF grant No. 1531492 (SeaWulf HPC cluster maintained by Research Computing and Cyberinfrastructure) and NSF grant No. 1919752 (Major Research Infrastructure program), respectively.

We would also like to thank our anonymous reviewers for their perceptive comments, which improved this work.

\bibliography{anthology,custom,nl}

\appendix

\section{Configuration Details for Experiments}
\label{sec:appendix-config}
\label{sec:config}

For all experiments we fine-tune Llama-3-8B on each of the five cross-validation folds with a batch size of 1 and no gradient accumulation steps. The AdamW optimizer is configured with a learning rate of 2e-5, weight decay of 0, and epsilon of 1e-8. As the cross-validation preparation does not contain a development set to conserve data, we train for a fixed 10 epochs. We configure LoRA with $\alpha$ of 16, dropout of 0.1, and $r$ of 64. Since $r$ is somewhat large, we observed slightly better results using rank-stabilization which scales adapters during forward passes by a factor of $\alpha/\sqrt{r}$, instead of the typical $\alpha/r$ \citep{kalajdzievski-2023-rank}. These parameters were arrived at through a run of hyperparameter tuning experiments.

\section{Supplementary Correlation Analysis} \label{sec:appendix-corrs}
This analysis was performed based our model (\secref{sec:fa-wiki}) output on the Wikipedia Talk Pages Corpus. Aside from \SNegT (e.g. thanking, commitments, accepting offers), \HPosT (e.g. criticism, insults, disapproval), and \Other (avoiding face altogether) the correlations have fairly low magnitude (absolute value less than 0.1).

\begin{table}[h]
\centering
\resizebox{\linewidth}{!}{
\begin{tabular}{lrr}
\toprule
 & \bfseries Politeness & \bfseries Impoliteness \\
\midrule
\bfseries \HNegT & 0.01 & 0.05 \\
\bfseries \HPosT & -0.11 & 0.18 \\
\bfseries \HNegR & -0.01 & 0.01 \\
\bfseries \HPosR & 0.03 & -0.04 \\
\midrule
\bfseries \SNegT & 0.31 & -0.25 \\
\bfseries \SPosT & 0.04 & -0.07 \\
\bfseries \SNegR & 0.00 & -0.01 \\
\bfseries \SPosR & -0.01 & -0.01 \\
\midrule
\bfseries None & -0.17 & 0.06 \\
\bottomrule
\end{tabular}}
\caption{Pearson's correlation coefficients between politeness scores and face acts.
}
\label{tab:fa-pp-corr}
\end{table}

\end{document}